\documentclass[letterpaper]{article} 
\usepackage{aaai2026}  
\usepackage{times}  
\usepackage{helvet}  
\usepackage{courier}  
\usepackage[hyphens]{url}  
\usepackage{graphicx} 
\usepackage{natbib}  
\usepackage{caption} 
\usepackage{algorithm}
\usepackage{algorithmic}
\usepackage{amsmath,amsfonts,amssymb}
\usepackage{booktabs}
\usepackage{multirow}
\usepackage{pifont}
\usepackage{boldline}
\usepackage{array}
\usepackage{graphicx}
\usepackage{arydshln}

\usepackage{newfloat}
\usepackage{listings}
\DeclareCaptionStyle{ruled}{labelfont=normalfont,labelsep=colon,strut=off} 
\floatstyle{ruled}
\newfloat{listing}{tb}{lst}{}
\floatname{listing}{Listing}
\title{3DTeethSAM: Taming SAM2 for 3D Teeth Segmentation}
\author{
    Zhiguo Lu\textsuperscript{\rm 1}, 
    Jianwen Lou\textsuperscript{\rm 1}\thanks{Corresponding author.},
    Mingjun Ma\textsuperscript{\rm 2},
    Hairong Jin\textsuperscript{\rm 3},
    Youyi Zheng\textsuperscript{\rm 3},
    Kun Zhou\textsuperscript{\rm 3}
}
\affiliations{

    \textsuperscript{\rm 1}School of Software Technology, Zhejiang University\\
    \textsuperscript{\rm 2}College of Intelligence and Computing, Tianjin University\\
    \textsuperscript{\rm 3}State Key Lab of CAD\&CG, Zhejiang University\\
    \{lzg, jianwen.lou\}@zju.edu.cn

}

\usepackage{bibentry}

\begin{document}

\maketitle

\begin{abstract}
3D teeth segmentation, involving the localization of tooth instances and their semantic categorization in 3D dental models, is a critical yet challenging task in digital dentistry due to the complexity of real-world dentition. In this paper, we propose 3DTeethSAM, an adaptation of the Segment Anything Model 2 (SAM2) for 3D teeth segmentation. SAM2 is a pretrained foundation model for image and video segmentation, demonstrating a strong backbone in various downstream scenarios. To adapt SAM2 for 3D teeth data, we render images of 3D teeth models from predefined views, apply SAM2 for 2D segmentation, and reconstruct 3D results using 2D-3D projections. Since SAM2's performance depends on input prompts and its initial outputs often have deficiencies, and given its class-agnostic nature, we introduce three light-weight learnable modules: (1) a prompt embedding generator to derive prompt embeddings from image embeddings for accurate mask decoding, (2) a mask refiner to enhance SAM2's initial segmentation results, and (3) a mask classifier to categorize the generated masks. Additionally, we incorporate Deformable Global Attention Plugins (DGAP) into SAM2's image encoder. The DGAP enhances both the segmentation accuracy and the speed of the training process. Our method has been validated on the 3DTeethSeg benchmark, achieving an IoU of 91.90\% on high-resolution 3D teeth meshes, establishing a new state-of-the-art in the field.
\end{abstract}

\begin{links}
    \link{Code}{https://github.com/Crisitofy/3DTeethSAM}
\end{links}

\begin{figure*}[htbp]
  \centering
  \includegraphics[width=\textwidth]{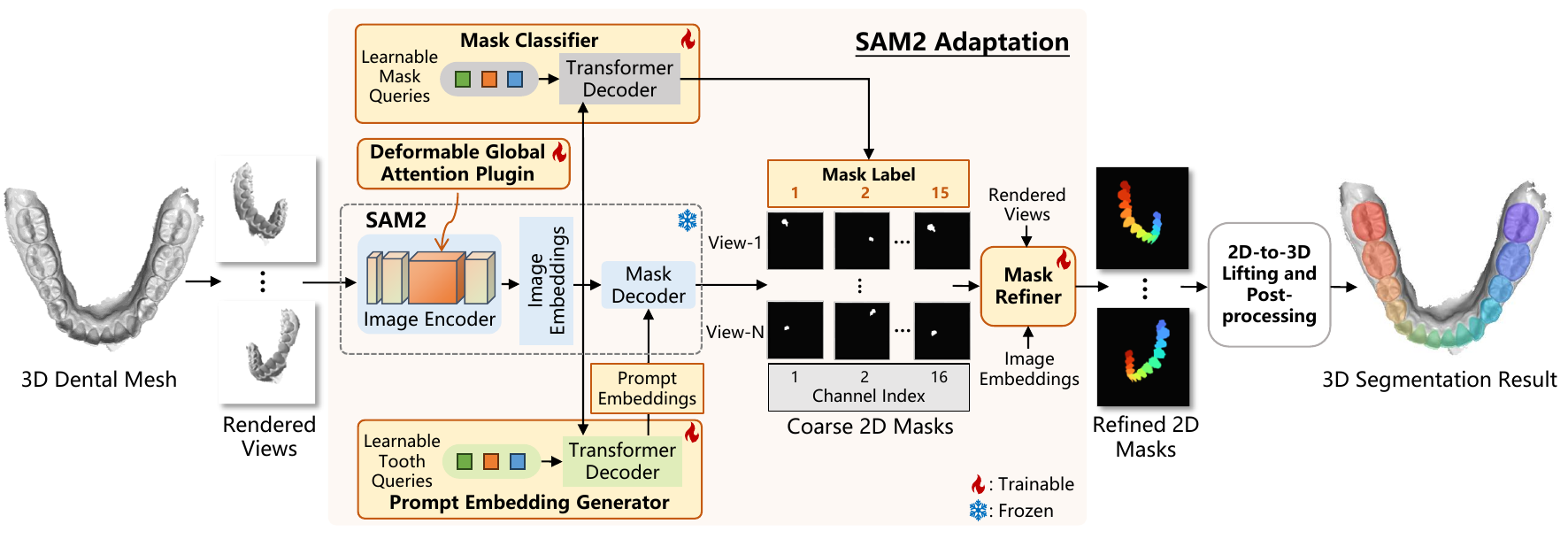}
  \caption{Pipeline of 3DTeethSAM. 3D dental models are rendered into 2D images from predefined views and processed by an adapted SAM2 for segmentation. The 2D results are then reconstructed in 3D. To enhance performance, three lightweight modules are introduced: (1) Prompt Embedding Generator for mask decoding, (2) Mask Refiner for segmentation improvement, and (3) Mask Classifier for semantic labeling. Additionally, the Deformable Global Attention Plugin (DGAP) refines feature sampling in SAM2's image encoder.}
  \label{fig:pipeline}
\end{figure*}

\section{Introduction}
Analyzing 3D dental models, which provide accurate and high-resolution representations of a patient’s oral anatomy, is a fundamental aspect of digital dentistry. This analysis enables precise diagnosis, effective treatment planning, and the creation of customized dental solutions. A crucial step in this process is the segmentation and classification of individual teeth, which facilitates further tasks such as orthodontic staging. However, 3D teeth segmentation remains a challenging problem due to two primary issues: (1) the complexity of real-world dentition, including anatomical variations and dental anomalies such as crowding and crooked teeth, and (2) the scarcity of labeled data.

Current methods for 3D teeth segmentation predominantly rely on deep neural networks (e.g., PointNet++\cite{qi2017pointnet++}, TSGCNet\cite{zhao2021two}, and TSRNet~\cite{jin2024tsrnet}) that operate directly on 3D dental models represented as meshes or point clouds. While these methods show promising results on low-resolution models, they struggle to scale to high-resolution 3D ones due to the limited capacity of the networks. In contrast, 2D vision foundation models, such as the Segment Anything Model 2 (SAM2)~\cite{ravi2024sam}, trained on billions of 2D masks, have demonstrated impressive image segmentation performance and generalization capabilities across various tasks. This success suggests a potential pathway for leveraging SAM2’s pretrained capabilities in 3D teeth segmentation. However, adapting SAM2 for this task poses significant challenges, including the inherent 2D-3D dimensionality mismatch, the need for manual prompts (such as points or boxes) to guide segmentation, and the difficulty of fine-tuning SAM2 effectively while preserving its pretrained weights.

In this paper, we propose 3DTeethSAM, a novel framework that adapts SAM2 for automatic 3D teeth segmentation. As shown in Figure~\ref{fig:pipeline}, our approach starts by rendering the 3D dental mesh into multiple 2D images from predefined viewing angles. These images are then processed by a customized version of SAM2 to generate segmentation masks. The 2D segmentation results are lifted back into 3D space using a voting strategy, which aggregates information from multiple views. To further refine the segmentation, we apply Graph Cut~\cite{boykov2001interactive} to correct boundary inaccuracies. A key innovation of our approach is the adaptation of SAM2 to 3D teeth segmentation through lightweight adapters and an image embedding enhancement scheme. These adapters, including a Prompt Embedding Generator for generating informative prompt embeddings, a Mask Refiner to improve the coarse segmentation, and a Mask Classifier to enable semantic label recognition, enhance SAM2's ability to handle 2D teeth images effectively. Additionally, we introduce the Deformable Global Attention Plugin (DGAP) into SAM2's image encoder. DGAP dynamically samples features during global attention, helping the model focus on the region of interest, i.e., the teeth. Extensive validation on the Teeth3DS benchmark demonstrates that 3DTeethSAM achieves a mean IoU of 91.90\%, outperforming existing methods and establishing an effective framework for adapting 2D foundation models to complex 3D segmentation tasks. The main contributions can be summarized as follows:
\begin{itemize}
    \item We propose 3DTeethSAM, a novel method that adapts the Segment Anything Model 2 (SAM2) for 3D teeth segmentation. The method includes three lightweight, learnable modules that enable auto-prompting and improve SAM2’s initial outputs, ensuring accurate segmentation and categorization of teeth. Additionally, we incorporate Deformable Global Attention Plugins (DGAP) to enhance both segmentation accuracy and training efficiency.
    \item We conduct extensive experiments that demonstrate the effectiveness of the proposed method and provide valuable insights into its key components.
\end{itemize}

\section{Related Work}
\subsection{3D Dental Segmentation}
The field of 3D dental segmentation has evolved from traditional geometry-based methods~\cite{yuan2010single, wu2014tooth, zou2015interactive} to deep learning approaches. Early techniques, which relied on curvature or watershed algorithms~\cite{li2007fast}, set initial benchmarks but struggled with the geometric complexity of real-world dentitions. The advent of deep learning, driven by networks like PointNet~\cite{qi2017pointnet} and DGCNN~\cite{wang2019dynamic}, led to the development of powerful, data-driven solutions. Today, state-of-the-art methods focus on specialized architectures, such as those that learn from mesh structures (e.g., MeshSegNet~\cite{lian2020deep}), incorporate transformers for geometric context (e.g., TsegFormer~\cite{xiong2023tsegformer}), or perform post-hoc refinement~\cite{jin2024tsrnet}. A common limitation among these advanced methods is their reliance on standalone networks trained from scratch on domain-specific data, which prevents them from leveraging the vast knowledge encapsulated in large-scale pre-trained models.

\subsection{Segment Anything Model}
The advent of 2D vision foundation models, particularly the Segment Anything Model (SAM)\cite{kirillov2023segment} and its successor SAM2~\cite{ravi2024sam}, has revolutionized computer vision with impressive zero-shot, prompt-based image segmentation. This has led to various adaptations, such as MedSAM\cite{ma2024segment}, which tailors SAM for medical imagery, and efforts~\cite{ke2023segment, fan2023stable} to enhance its adaptability to downstream tasks. While SAM2 has proven highly successful in 2D image segmentation, its application to 3D segmentation remains largely unexplored. Key challenges include the inherent 2D-3D dimensionality mismatch, reliance on manual prompts (e.g., points or boxes) for guidance, and difficulties in fine-tuning SAM2 while preserving its pretrained weights. Despite these challenges, applying 2D foundation models like SAM2 to 3D teeth segmentation holds great promise, a direction we explore in this study.

\section{Methodology}
We propose 3DTeethSAM, a novel framework that leverages the 2D vision foundation model SAM2~\cite{ravi2024sam} for 3D teeth segmentation. As illustrated in Figure~\ref{fig:pipeline}, our approach begins by rendering a 3D dental mesh into multiple images from predefined viewing angles. These images are then processed using a customized version of SAM2 to perform segmentation. The resulting 2D segmentation masks are lifted back into 3D space via an intuitive voting strategy, which aggregates information across different views. Finally, to refine the segmentation, we apply Graph Cut~\cite{boykov2001interactive} to correct issues like inaccuracies in the boundaries. At the heart of our approach is the adaptation of a pretrained SAM2 model to effectively handle teeth images. This is achieved by introducing learnable lightweight adapters and an image embedding enhancement scheme, while preserving SAM2’s pretrained weights. Specifically, the adapters include a Prompt Embedding Generator, which creates informative prompt embeddings from image features for mask decoding; a Mask Refiner, which optimizes the coarse segmentation results produced by SAM2; and a Mask Classifier, which enables SAM2 to be class-aware by recognizing semantic labels in the segmentation map. To extract an enhanced image embedding for the adapters, we integrate a key component into SAM2's image encoder: Deformable Global Attention Plugin (DGAP). DGAP dynamically samples features during global attention, helping the model focus on the region of interest, i.e., the teeth. In the following sections, we first introduce the preliminaries, including the basic concepts of 3D teeth segmentation and SAM2, and then describe our proposed method in detail.

\subsection{Preliminaries}
\subsubsection{3D Teeth Segmentation}
Both the upper and lower halves of an adult's jaw contain up to 16 teeth, with each tooth having a unique ID. The goal of 3D teeth segmentation is to assign a class label to each vertex in the 3D dental model, identifying one of the 16 teeth or the background. In this study, a 3D dental model is represented as a 3D mesh of either the upper or lower teeth. It is denoted as \(\mathcal{M}(P, F)\), where \(P=\{p_n\in \mathbb{R}^3,n=1,2,...,N\}\) is the set of 3D points, and \(F=\{(p_i,p_j,p_k)_m,m=1,2,...,M\}\) is the set of triangles that define the connectivity between points.

\subsubsection{SAM2}
Segment Anything Model 2 (SAM2) is a groundbreaking vision foundation model designed for promptable segmentation in images and videos. It consists of three main components: an image encoder, a prompt encoder, and a mask decoder. The image encoder, based on the Vision Transformer (ViT) architecture, processes the input image and generates an embedding that captures high-level visual features. The prompt encoder embeds prompts, such as points, boxes, or masks. The mask decoder combines the above two information sources to predict segmentation masks.

\subsection{Multi-View Teeth Image Rendering}
Given a 3D dental mesh \(\mathcal{M}\), we render it into multi-view images to align with SAM2 for segmentation. This process is designed to comprehensively capture both structural and surface information of the 3D dental mesh from a set of carefully chosen viewpoints, ensuring that critical features for segmentation are visible in at least one rendered view. Specifically, each 3D dental mesh is first normalized by translating its center to the origin of a uniform coordinate system and rotating it so that the dental crown is oriented upwards. The transformed mesh is then rendered into 512x512 RGB images from a set of fixed camera viewpoints (see Figure~\ref{fig:pipeline}), including a frontal view, a back view, and several side views. The rendering process is formulated as follows:
\begin{equation}
I_v = \Pi_v(\mathcal{M})
\end{equation}
where \(I_v\in \mathbb{R}^\text{3x512x512}\) represents the rendered teeth image, and \(\Pi_v\) denotes the camera projection from the viewing angle \(v\). The number of views is empirically set to balance segmentation accuracy and computational complexity. 

Using the above equation, we can also render a teeth mask map \(Mask_v\in \mathbb{R}^\text{16x512x512}\) from the 3D dental mesh based on point-wise segmentation labels. The mask map has 16 channels, each corresponding to one of the 16 teeth. Its pixel values are normalized to the range 0 to 1 along each channel. The value indicates the probability that a pixel belongs to the tooth category associated with that particular channel.

\subsection{Teeth Image Segmentation via SAM2 Adaptation}
Trained on a large-scale dataset of natural images, SAM2 demonstrates strong zero-shot learning performance across a wide range of downstream tasks. However, applying the pretrained SAM2 model to teeth images is not straightforward. There are three key challenges: first, SAM2 requires input prompts for segmentation; second, the original segmentation results produced by SAM2 show noticeable deficiencies; and third, SAM2 is class-agnostic. These issues prevent the vanilla SAM2 model from meeting the automation and precision requirements for teeth image segmentation. To leverage the pre-learned knowledge of SAM2 and adapt it for teeth image segmentation in a fully automatic and parameter-efficient manner, we introduce three lightweight SAM2 adapters: a Prompt Embedding Generator, a Mask Refiner, and a Mask Classifier. To further enhance the model’s adaptation, we propose integrating the Deformable Global Attention Plugin (DGAP) into SAM2's image encoder, which improves feature extraction for the adapters. During training, we keep SAM2's pretrained weights frozen, while only optimizing those of the adapters and DGAP.

\subsubsection{Prompt Embedding Generator}
SAM2 uses prompts such as points, boxes, and masks to identify the target image region for segmentation. The type and location of the prompts are critical for accurate segmentation, but they are often difficult to determine. Additionally, when multiple segmentation targets are present in an image and have an inherent structure (e.g., teeth arranged in a canonical pattern), it becomes essential to model the positional relationships between the prompts. To address these challenges, we propose a trainable generator capable of predicting prompt embeddings directly from image features. Specifically, inspired by DETR~\cite{carion2020end}, we employ a Transformer decoder for this generation process. The decoder follows the architecture of the original Transformer decoder introduced in \cite{vaswani2017attention}. It accepts randomly initialized query vectors as input and transforms them into prompt embeddings for tooth instances using multi-layer self-attention and cross-attention. The self-attention module models pairwise relationships between the queries, while the cross-attention module aligns the queries with the image features. This approach allows the model to reason about all prompt embeddings simultaneously, leveraging both the relationships between tooth instances and the broader image context. Given that a teeth image can contain up to 16 teeth, we set the number of query vectors to 16. To handle cases with missing teeth, we also learn a confidence score from each prompt embedding, using a fully connected layer and a Sigmoid function. This score indicates the validity of each prompt embedding, with values ranging from 0 to 1. A higher score represents a greater probability that the corresponding tooth instance exists.

\subsubsection{Mask Refiner}
Using the generated prompt embeddings and the teeth image embedding output by SAM2's image encoder, we can create a 16-channel mask map with SAM2's mask decoder. Each channel in this map corresponds to the segmentation mask of a specific tooth. However, these masks may not precisely localize tooth instances, particularly along the boundaries, due to SAM2's general-domain pre-training. To address the limitations of the coarse masks, we introduce a mask refiner, a specialized convolutional neural network designed to enhance boundary precision. The refiner processes three key inputs:
\begin{itemize}
    \item Teeth image: Provides rich, low-level shape and texture details crucial for precise boundary delineation.
    \item Coarse mask map: Offers strong spatial priors on the tooth instances’ location and shape.
    \item SAM2's image embedding: Captures high-level semantic context.
\end{itemize}
The backbone of the refiner is based on a UNet~\cite{ronneberger2015u} architecture, which is commonly used for image segmentation. In the contracting path of the UNet, each convolutional layer contains three main streams, with each stream transforming a specific input into latent representations. These latent features are then concatenated and passed to the subsequent convolutional layer and the expansive path of the UNet via a skip connection.

\subsubsection{Mask Classifier} 
The original SAM2 model does not support identifying the semantic category of each segmentation mask. To address this limitation, a straightforward approach would be to bind the channels of the mask map \(Mask_v\in \mathbb{R}^\text{16x512x512}\) to 16 tooth IDs in a one-to-one manner. In this setup, each channel would correspond to a specific tooth ID, allowing the model to infer the tooth identity directly from the channel index. However, our experiments show that this naive method is prone to channel-to-ID mismatches, often assigning a tooth mask to the wrong channel, especially in cases with missing teeth. To overcome this issue, we introduce a mask classifier that explicitly identifies the tooth ID associated with each channel in the mask map. The classification process resembles the prompt embedding generation in SAM2: it requires capturing both the spatial correlation between teeth (implied by their natural arrangement) and the image context. To this end, we adopt the same architecture as the prompt embedding generator, using a Transformer decoder ~\cite{vaswani2017attention} to transform 16 randomly initialized query vectors (each representing a potential tooth instance) into 16 class probability vectors, based on image features. The only architectural difference is in the classifier’s final layers, which consist of a multi-layer perceptron (MLP) followed by a softmax activation. To handle missing teeth, we extend each class probability vector with an additional dimension representing the background. This allows the model to classify absent tooth instances as background, ensuring robustness in cases with incomplete dental data.  

\begin{figure}[t] 
  \centering
  \includegraphics[width=\linewidth]{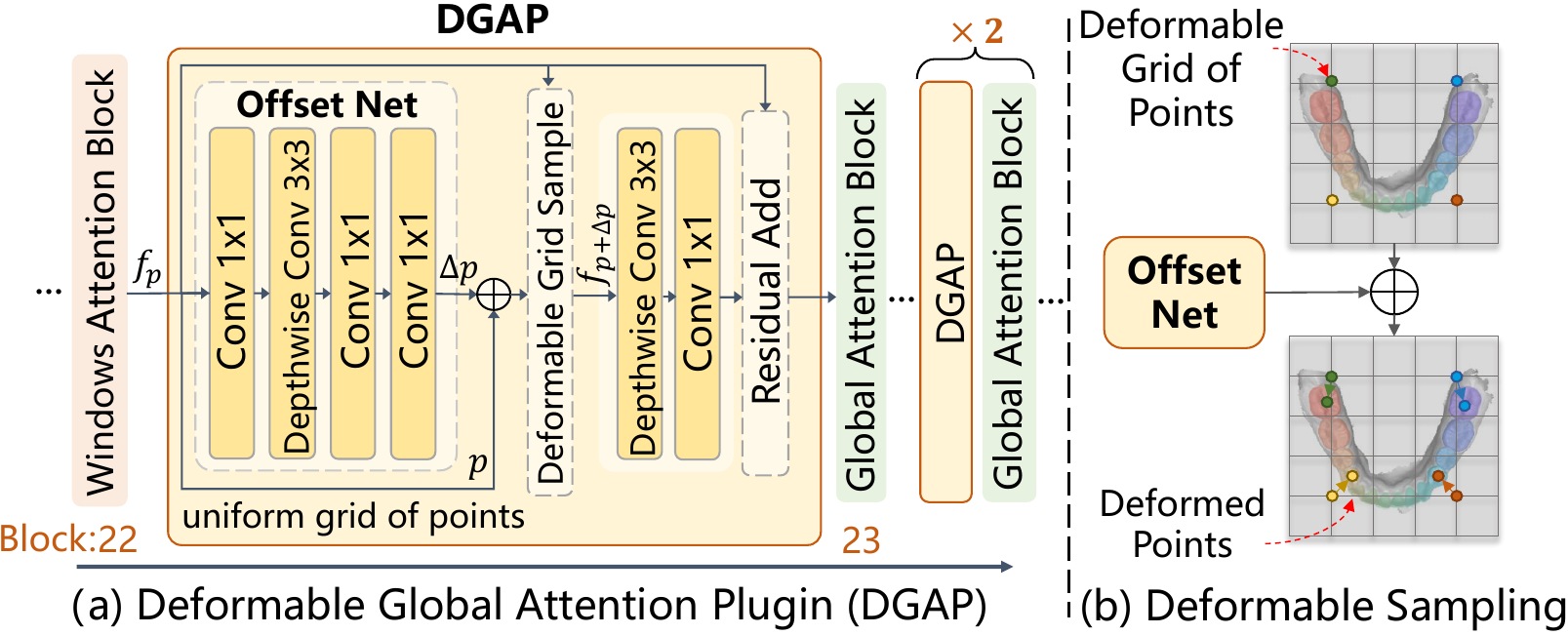}
    \caption{Illustration of the proposed Deformable Global Attention Plugin (DGAP). (a) Architecture of the DGAP module. (b) Deformable sampling grid based on the Offset Net.}
    \label{fig:dgap}
\end{figure}

\begin{figure}[t]
    \centering
    \includegraphics[width=0.8\linewidth]{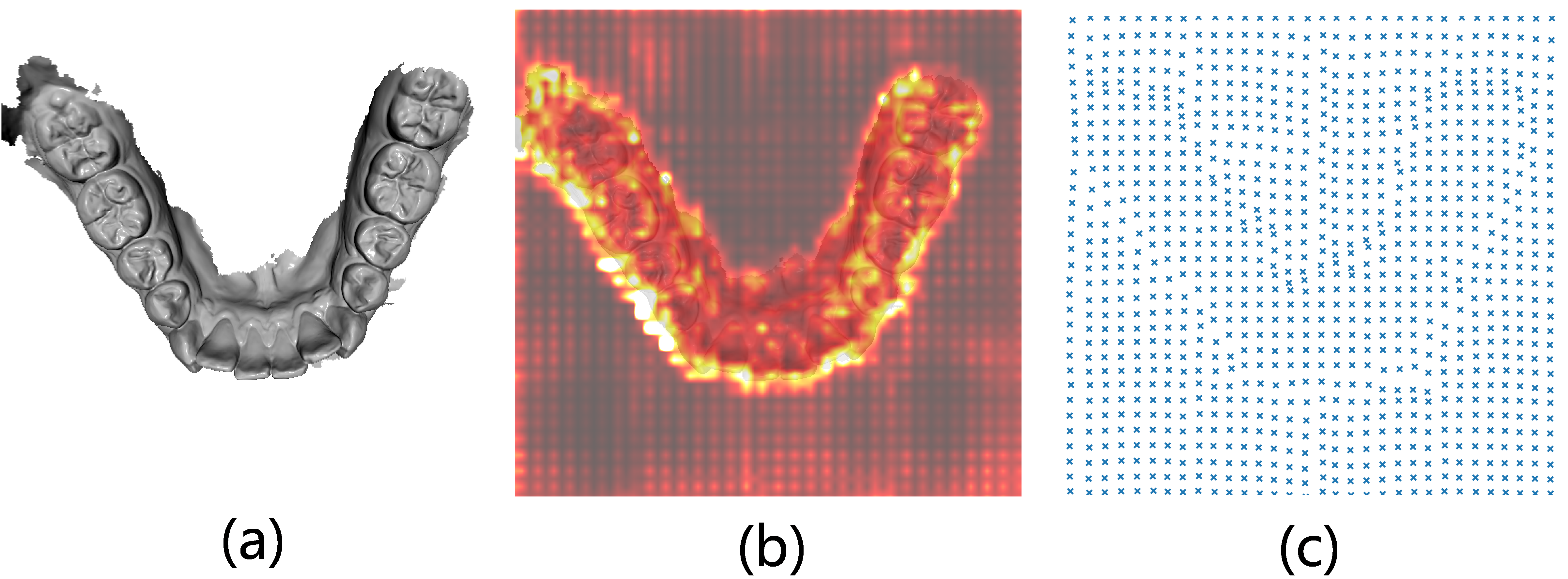}
    \caption{Effect of the Deformable Global Attention Plugin (DGAP). (a) Rendered teeth image. (b) Attention heatmap after applying DGAP. (c) Deformed sampling points that align with the shape of the teeth.}
    \label{fig:dgap_vis}
\end{figure}

\subsubsection{Deformable Global Attention Plugin}
The above three adapters all require an image embedding as input for prediction. Although the feature pyramid produced by SAM2's image encoder offers a handy image embedding for the adapters, we argue that such a feature representation is designed for SAM2's mask decoding, not purely fitting with the adaptation process. A simple solution to this problem would be full-parameter tuning SAM2's image encoder during adaptation, which however is very costly. As a remedy, we propose to integrate a lightweight Deformable Global Attention Plugin into SAM2's image encoder for capturing informative features for adapters in a parameter-efficient way (see Figure~\ref{fig:dgap}). The plugin is inspired by the deformable attention mechanism introduced in~\cite{xia2022vision}. Deformable attention uses a dynamic feature map to generate key and value embeddings. It learns an offset network that predicts a set of offsets from the query embedding, which are then used to deform a grid of reference points on the feature map. This process enables dynamic feature extraction during the attention operation. The deformed points are shown to be concentrated around the target region in the image, which leads to more informative feature extraction. We incorporate deformable attention into SAM2's image encoder with a systematic adaptation as follows:
\begin{itemize}
    \item \textbf{Integration into Global Attention Block:} Deformable attention is integrated into each global attention block in the 3rd stage of SAM2's image encoder, specifically in the Hiera trunk. The Hiera trunk consists of four main stages, each containing multiple attention blocks. As the feature map down-samples across the stages, it becomes progressively more abstract. The 3rd stage is crucial as it contains the majority of attention blocks, which are essential for learning image embeddings. Therefore, we focus on deformable attention in this stage to extract the most meaningful features. Additionally, we choose the global attention block to maintain a broad receptive field. 
    \item \textbf{Embedding Prediction from Deformed Feature Map:} We predict not only the key and value embeddings but also the query embedding from the deformed feature map. This approach offers more flexibility than the standard deformable attention method. It also results in a plug-and-play module, where the offset network and dynamic sampling can be added before the global attention operation without altering the internal implementation.
    \item \textbf{Skip Connection for Feature Map Combination:} We combine the deformed and undeformed feature maps using a skip connection, allowing the model to leverage both, which results in more robust feature learning.
\end{itemize} 

\begin{figure*}[t]
    \centering
    \includegraphics[width=\textwidth]{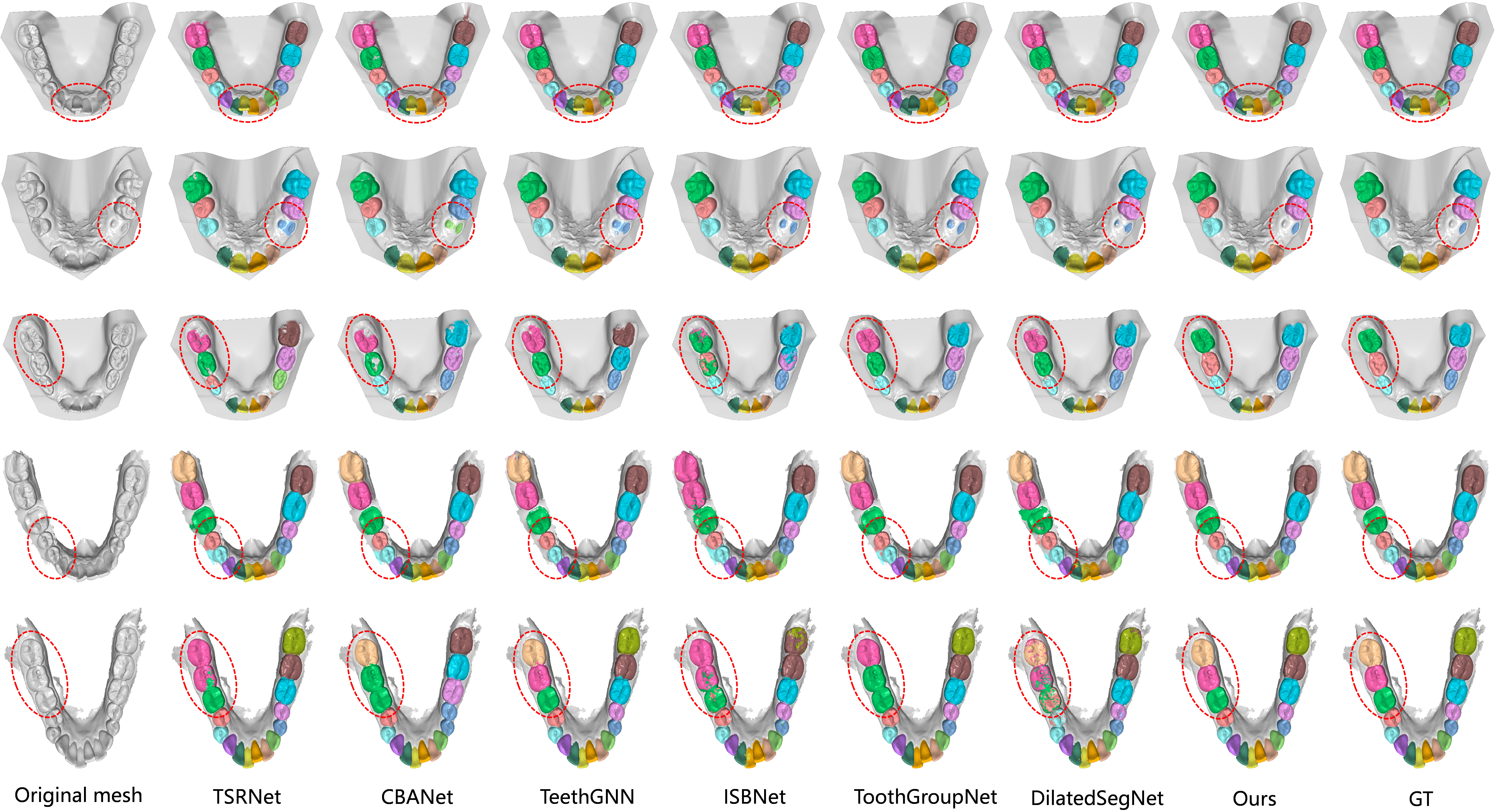}
    \caption{Visual comparison of different methods.}
    \label{fig:lower_comparison}
\end{figure*}

\subsubsection{Training Losses}
\label{sec:loss}
Our training employs a composite loss function to address SAM2's adaptation challenges: automated prompting, semantic classification, and boundary refinement. We use the Hungarian algorithm~\cite{kuhn1955hungarian} for optimal one-to-one assignment between predicted queries and ground-truth annotations, enabling instance-aware supervision.

The total loss function jointly optimizes our three lightweight adapters:
\begin{equation}
\label{eq:total_loss}
    L_{\text{total}} = \lambda_{\text{MC}}L_{\text{MC}} + \lambda_{\text{PEG}}L_{\text{PEG}} + \lambda_{\text{MR}}L_{\text{MR}}
\end{equation}
where $\lambda_{\text{MC}}$, $\lambda_{\text{PEG}}$, and $\lambda_{\text{MR}}$ are hyperparameter weights that balance the contribution of each adapter.

\textbf{Mask Classifier Loss ($L_{\text{MC}}$):} The classifier addresses SAM2's class-agnostic limitation by learning to distinguish between different tooth types. We apply Cross-Entropy loss over 17 classes (16 teeth + background) to matched query-target pairs:
\begin{equation}
L_{\text{MC}} = -\sum_{i \in \mathcal{M}} \log p_i(c_i^*)
\end{equation}
where $\mathcal{M}$ denotes matched pairs, $c_i^*$ is the ground-truth class, and $p_i(c_i^*)$ is the predicted probability. 

\textbf{PEG Loss ($L_{\text{PEG}}$):} The PEG must generate high-quality prompt embeddings that produce accurate initial masks while handling missing teeth scenarios. Its loss combines three complementary terms:
\begin{equation}
\label{eq:peg_loss}
L_{\text{PEG}} = \lambda_{\text{bce}}L_{\text{bce}} + \lambda_{\text{dice}}L_{\text{dice}} + \lambda_{\text{conf}}L_{\text{conf}}
\end{equation}
Here, $L_{\text{bce}}$ provides pixel-wise supervision through binary cross-entropy, $L_{\text{dice}}$ optimizes region-level overlap to handle class imbalance, and $L_{\text{conf}}$ trains confidence scores to predict tooth presence, enabling robust handling of incomplete dental arches.

\textbf{Mask Refiner Loss ($L_{\text{MR}}$):} The refiner transforms coarse masks into precise segmentations with sharp boundaries—critical for clinical applications. Its multi-objective loss ensures both semantic accuracy and boundary precision:
\begin{equation}
\label{eq:mr_loss}
L_{\text{MR}} = \lambda_{\text{ce}}L_{\text{ce}} + \lambda_{\text{dice}}L_{\text{dice}} + \lambda_{\text{boundary}}L_{\text{boundary}}
\end{equation}
where $L_{\text{ce}}$ applies multi-class Cross-Entropy over all 17 classes, $L_{\text{dice}}$ maximizes regional overlap, and $L_{\text{boundary}}$ measures L1 distance between spatial gradients of predicted and ground-truth masks using Sobel filters, encouraging sharp tooth-gingiva boundaries essential for treatment planning.

\subsection{2D-to-3D Segmentation Lifting and Postprocessing}
After obtaining the 2D segmentation results using the adapted SAM2 model, we lift these results into 3D space. Specifically, we invert the projection matrix applied during multi-view image rendering and assign the segmentation label of each image pixel to its corresponding 3D mesh vertex. In cases where multiple 2D segmentation labels are associated with a single mesh vertex, we select the label that appears most frequently across all rendered views. As in previous studies, we apply the well-known Graph Cut approach~\cite{boykov2001interactive} to further refine the 3D segmentation by filling holes and improving boundary precision.

\section{Experiments}
\subsubsection{Datasets}
We conduct comprehensive experiments on the Teeth3DS benchmark, the most challenging publicly available dataset for 3D dental segmentation. The dataset comprises 1,800 high-resolution intraoral 3D scans from 900 patients, with both upper and lower dental arches included. Following the FDI dental notation system, each dental arch contains up to 16 tooth instances plus gingival background, resulting in 17 semantic classes. We strictly adhere to the official train/test split (1,200/600 scans) to ensure fair comparison with prior methods.

\subsubsection{Implementation Details}
All experiments are conducted using PyTorch on NVIDIA RTX 4090 GPUs. We employ SAM2 (Hiera-L) as our foundation model backbone, keeping its pre-trained weights frozen. The network processes original-resolution 3D meshes without downsampling, with multi-view renderings dynamically generated at 512×512 resolution.

Training adopts an end-to-end scheme with AdamW optimizer (learning rate: 2e-4, cosine annealing schedule with 5-epoch warm-up). Models are trained for 100 epochs with batch size 4, utilizing mixed-precision training for efficiency. The loss weights are empirically set as: $\lambda_{\text{MC}}=1.0$, $\lambda_{\text{PEG}}=1.0$, $\lambda_{\text{MR}}=2.0$, with sub-loss weights following standard practices.

\subsubsection{Evaluation Metrics}
The evaluation encompasses four complementary metrics: overall accuracy (OA), tooth-wise mIoU (T-mIoU), boundary IoU (B-IoU), and Dice score. OA measures per-vertex classification correctness across the entire dental mesh. T-mIoU computes the mean IoU across all individual tooth instances, providing an instance-level assessment of segmentation quality. The Dice score offers a global measure of region overlap between predicted and ground-truth masks. B-IoU specifically evaluates boundary precision by focusing on vertices near inter-tooth boundaries, where a vertex is considered a boundary point if there exist vertices with different labels within its k-neighbourhood (k = 10). This boundary-focused metric is particularly crucial for accurate crown–gingiva delineation in clinical applications. 

\subsubsection{Comparison on Teeth3DS}
\label{sec:quantitative_results}

Table~\ref{tab:segmentation} presents quantitative comparisons against 11 state-of-the-art methods spanning different paradigms: point cloud networks (PointNet++, DGCNN), mesh-based approaches (MeshSegNet, iMeshSegNet), graph neural networks (TeethGNN, TSGCNet), and recent advanced methods (TSRNet, ToothGroupNet). 

3DTeethSAM achieves state-of-the-art performance across all metrics: 95.48\% OA, 91.90\% T-mIoU, 70.05\% B-IoU, and 94.33\% Dice. Notably, our method demonstrates substantial improvements over the best-performing prior method (ToothGroupNet) by +1.74\% T-mIoU and +0.75\% B-IoU, establishing new benchmarks while using fewer trainable parameters than methods trained from scratch.

Looking at the eight symmetric tooth groups in the rightmost columns, 3DTeethSAM consistently achieves the highest mIoU for every group, with particularly strong performance on rare and challenging categories. For instance, it attains 83.29\% mIoU on wisdom teeth (T$_{8/16}$), significantly outperforming ToothGroupNet’s 68.2\%. This 15.09-point improvement highlights the benefits of leveraging large-scale 2D pretraining, demonstrating how foundation models can effectively mitigate data scarcity issues that often hinder specialized 3D architectures.

Figure~\ref{fig:lower_comparison} illustrates our method's robustness across diverse clinical scenarios: dental crowding, tooth eruption, edentulous regions, gingival complexity, and complete arch segmentation. Unlike prior methods that suffer from boundary ambiguity or class confusion, 3DTeethSAM maintains precise inter-tooth boundaries and avoids common failure modes such as gingival leakage or adjacent tooth merging.

\begin{table*}[tb]
  \centering
  \small 
  \setlength{\tabcolsep}{3pt} 
    \begin{tabular}{l cccc cccccccc}
      \toprule
      \textbf{Method} & \textbf{OA} & \textbf{T\textsubscript{all}} & \textbf{B\textsubscript{all}} & \textbf{Dice} & \textbf{T\textsubscript{1/9}} & \textbf{T\textsubscript{2/10}} & \textbf{T\textsubscript{3/11}} & \textbf{T\textsubscript{4/12}} & \textbf{T\textsubscript{5/13}} & \textbf{T\textsubscript{6/14}} & \textbf{T\textsubscript{7/15}} & \textbf{T\textsubscript{8/16}} \\
      \midrule
      iMeshSegNet~\cite{wu2022two} & 82.45 & 67.65 & 26.31 & 77.58 & 71.63 & 70.90 & 68.20 & 71.67 & 66.98 & 67.66 & 55.65 & 00.00 \\
      TSegNet~\cite{cui2021tsegnet} & 78.22 & 59.81 & 28.00 & 67.35 & 57.05 & 62.89 & 69.03 & 58.74 & 52.23 & 61.45 & 66.02 & 00.00 \\
      TeethGNN~\cite{zheng2022teethgnn} & 90.96 & 83.89 & 48.49 & 88.46 & 86.58 & 86.31 & 86.56 & 87.69 & 82.05 & 79.21 & 82.89 & 66.54 \\
      TSRNet~\cite{jin2024tsrnet} & 92.87 & 86.56 & 51.11 & 90.91 & 88.20 & 87.83 & 87.35 & 87.65 & 86.57 & 86.69 & 83.74 & 70.82 \\
      ToothGroupNet~\cite{ben20233dteethseg} & 95.19 & 90.16 & 69.30 & 92.88 & 92.19 & 92.30 & 92.65 & 93.44 & 87.74 & 88.84 & 84.82 & 68.2 \\ 
      TSGCNet~\cite{zhang2021tsgcnet} & 89.84 & 79.79 & 36.98 & 86.73 & 82.63 & 81.59 & 82.82 & 82.75 & 80.04 & 80.60 & 69.27 & 34.23 \\
      IsbNet~\cite{ngo2023isbnet} & 91.53 & 81.01 & 39.61 & 87.65 & 68.61 & 80.62 & 84.06 & 86.44 & 84.61 & 85.79 & 80.44 & 26.75 \\
      DGCNN~\cite{wang2019dynamic} & 90.21 & 80.08 & 28.60 & 87.27 & 80.85 & 80.31 & 81.12 & 82.67 & 79.86 & 81.53 & 75.45 & 52.42 \\
      CBAnet~\cite{jin2025learning} & 92.81 & 86.77 & 50.81 & 91.16 & 88.62 & 87.94 & 88.17 & 88.41 & 86.39 & 86.12 & 83.05 & 76.88 \\
      PT~\cite{zhao2021point} & 86.15 & 71.52 & 29.22 & 78.84 & 71.26 & 72.15 & 72.08 & 72.50 & 69.55 & 73.27 & 74.88 & 2.27 \\
      DilatedSegNet~\cite{krenmayr2024dilatedtoothsegnet} & 93.40 & 86.55 & 51.57 & 91.26 & 86.49 & 86.46 & 87.33 & 88.61 & 86.83 & 88.09 & 83.88 & 62.49 \\
      \midrule
      \textbf{Ours} & \textbf{95.48} & \textbf{91.90} & \textbf{70.05} & \textbf{94.33} & \textbf{93.28} & \textbf{93.36} & \textbf{93.16} & \textbf{93.79} & \textbf{91.65} & \textbf{90.73} & \textbf{89.10} & \textbf{83.29} \\
      \bottomrule
  \end{tabular}
  \caption{Comparison of segmentation performance. T\textsubscript{all} and B\textsubscript{all} are abbreviations for the mean IoU over all tooth classes (T-mIoU) and the boundary (B-IoU), respectively. Columns T\textsubscript{1/9} through T\textsubscript{8/16} present the per-tooth T-mIoU scores, with labels grouped for symmetric positions (e.g., T1 and T9).}
  \label{tab:segmentation}
\end{table*}

\begin{table}[htbp]
\centering
\setlength{\tabcolsep}{4pt}
\begin{tabular}{lcccc}
\toprule
\textbf{Setting} & \textbf{OA}↑ & \textbf{T-mIoU}↑ & \textbf{B-IoU}↑ & \textbf{Dice}↑ \\
\midrule
Full Model             & \textbf{95.48} & \textbf{91.90} & \textbf{70.05} & \textbf{94.33} \\
w/o DGAP               & 94.87 & 90.61 & 66.64 & 93.45 \\
w/o PEG                & 76.52 & 52.46 & 28.17 & 58.88 \\
w/o Mask Refiner       & 95.13 & 91.10 & 68.43 & 93.67 \\
w/o Mask Classifier    & 95.21 & 91.31 & 67.56 & 93.98 \\
\bottomrule
\end{tabular}
\caption{Ablation study on each proposed component. Removing any module results in a noticeable drop across all metrics, confirming the contribution of each module.}
\label{tab:component_ablation}
\end{table}

\begin{figure}[t]
    \centering
    \includegraphics[width=0.7\linewidth]{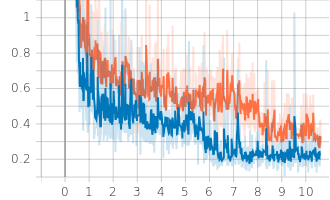}
    \caption{Training loss curves of our method on the Teeth3DS dataset. The blue curve (with DGAP) converges faster and reaches a lower final value compared to the orange curve (without DGAP), demonstrating the efficiency of the proposed DGAP.}
    \label{fig:train_loss}
\end{figure}

\subsubsection{Ablation Studies}
\label{subsec:ablation_studies}

To quantify the contribution of each module, we conduct an ablation study on the four key components of 3DTeethSAM. Starting from the full model, we progressively remove each component to evaluate its individual impact. The results are summarized in Table~\ref{tab:component_ablation}.

\textbf{Prompt Embedding Generator (PEG):} To evaluate PEG's contribution, we replace it with manual prompts derived from ground-truth mask center points, simulating an idealized manual prompting scenario. Even with this oracle-level prompting, performance drops dramatically by 39.44\% T-mIoU (91.90\%→52.46\%), demonstrating that PEG's automated, instance-aware prompt generation is fundamentally superior to traditional point-based prompting. This massive degradation confirms that our learned prompt embeddings capture complex spatial relationships and contextual information that simple point prompts cannot provide.
 
\textbf{Deformable Global Attention Plugin (DGAP):} Removing DGAP degrades T-mIoU by 1.29\% (91.90\%→90.61\%) and B-IoU by 3.41\%, confirming the benefit of morphology-aware attention for dental structures. Beyond performance gains, Figure~\ref{fig:train_loss} illustrates that DGAP significantly accelerates training convergence, providing both accuracy and efficiency benefits through deformable feature sampling.

\textbf{Mask Refiner:} Removing the refiner reduces T-mIoU by 0.80\% while degrading B-IoU by 1.62\%, indicating its role in both overall segmentation quality and boundary precision. 

\begin{figure}[t]
    \centering
    \includegraphics[width=\linewidth]{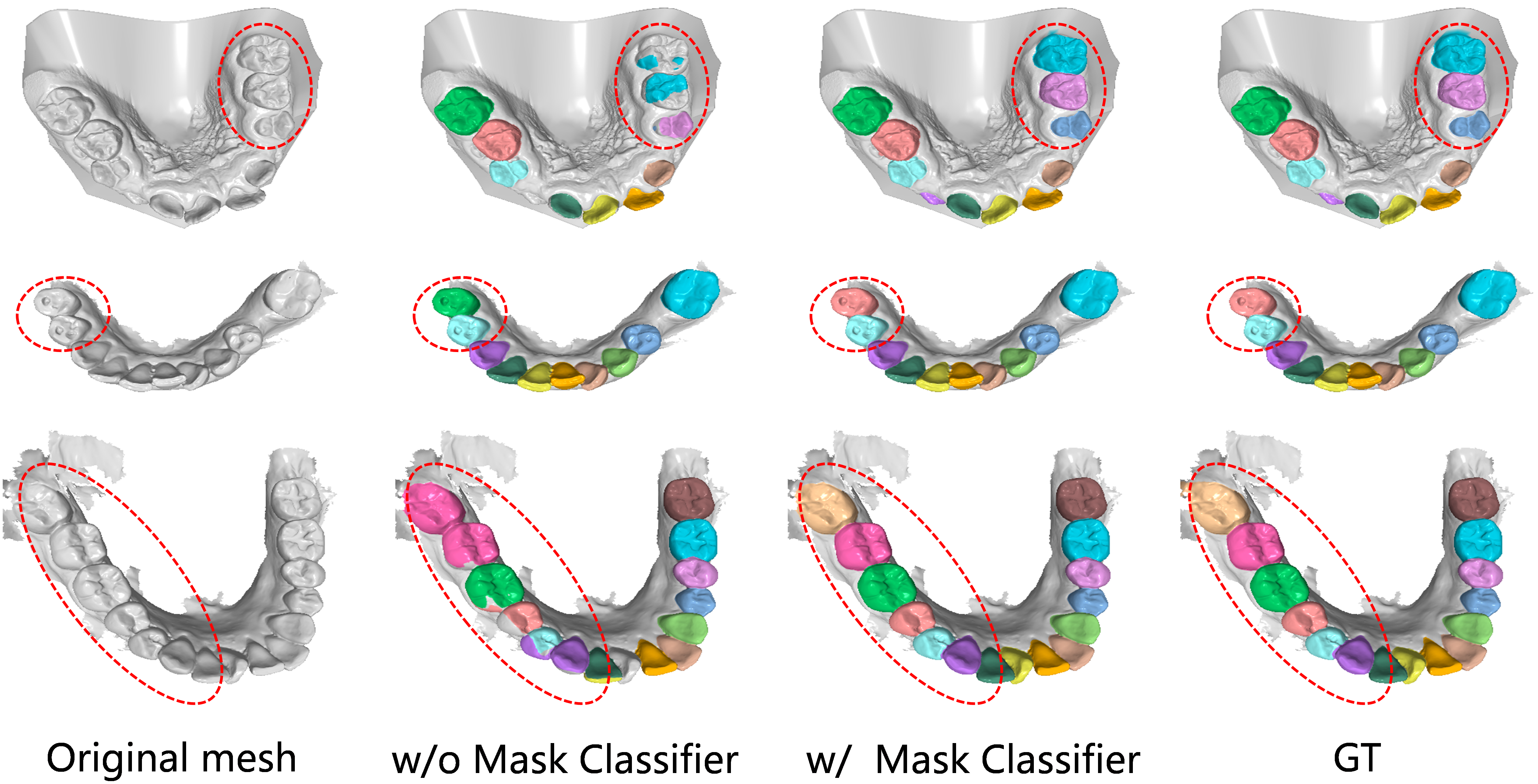}
    \caption{Visual results of the ablation study on the classifier module. The red ellipses highlight regions where the classifier corrects tooth category assignments, demonstrating its effectiveness in disambiguating adjacent teeth.}
    \label{fig:classifier_ablation}
\end{figure}

\textbf{Mask Classifier:} Disabling the classifier decreases T-mIoU by 0.59\% and B-IoU by 2.49\%, demonstrating that explicit semantic reasoning provides measurable benefits. Figure~\ref{fig:classifier_ablation} shows qualitative examples where the classifier correctly resolves tooth category assignments, particularly for morphologically similar adjacent teeth.

These findings confirm that each proposed component contributes complementary capabilities (DGAP for enhanced feature extraction, PEG for automated prompting, Mask Refiner for boundary refinement, and Mask Classifier for semantic disambiguation), which are essential for achieving SOTA 3D teeth segmentation performance.

\section{Conclusion}
In this paper, we present 3DTeethSAM, a novel framework that adapts SAM2 for automatic 3D teeth segmentation. Our approach involves rendering 3D dental meshes into 2D images, processing them with a customized SAM2 to generate segmentation masks, and reconstructing the results in 3D using a voting strategy. Key innovations include lightweight adapters for prompt generation, mask refinement, and semantic classification, as well as the introduction of the Deformable Global Attention Plugin (DGAP) to enhance feature sampling. Extensive validation on the Teeth3DS benchmark shows that 3DTeethSAM surpasses existing methods and demonstrating the effectiveness of adapting 2D foundation models for 3D segmentation tasks.

\bibliography{aaai2026}

@article{zhao2021two,
  title={Two-stream graph convolutional network for intra-oral scanner image segmentation},
  author={Zhao, Yue and Zhang, Lingming and Liu, Yang and Meng, Deyu and Cui, Zhiming and Gao, Chenqiang and Gao, Xinbo and Lian, Chunfeng and Shen, Dinggang},
  journal={IEEE Transactions on Medical Imaging},
  volume={41},
  number={4},
  pages={826--835},
  year={2021},
  publisher={IEEE}
}

@article{ben20233dteethseg,
  title={3DTeethSeg'22: 3D Teeth Scan Segmentation and Labeling Challenge},
  author={Ben-Hamadou, Achraf and Smaoui, Oussama and Rekik, Ahmed and Pujades, Sergi and Boyer, Edmond and Lim, Hoyeon and Kim, Minchang and Lee, Minkyung and Chung, Minyoung and Shin, Yeong-Gil and others},
  journal={arXiv preprint arXiv:2305.18277},
  year={2023}
}

@article{wu2014tooth,
  title={Tooth segmentation on dental meshes using morphologic skeleton},
  author={Wu, Kan and Chen, Li and Li, Jing and Zhou, Yanheng},
  journal={Computers \& Graphics},
  volume={38},
  pages={199--211},
  year={2014},
  publisher={Elsevier}
}

@article{yuan2010single,
  title={Single-tooth modeling for 3D dental model},
  author={Yuan, Tianran and Liao, Wenhe and Dai, Ning and Cheng, Xiaosheng and Yu, Qing},
  journal={International journal of biomedical imaging},
  volume={2010},
  number={1},
  pages={535329},
  year={2010},
  publisher={Wiley Online Library}
}

@article{zou2015interactive,
  title={Interactive tooth partition of dental mesh base on tooth-target harmonic field},
  author={Zou, Bei-ji and Liu, Shi-jian and Liao, Sheng-hui and Ding, Xi and Liang, Ye},
  journal={Computers in biology and medicine},
  volume={56},
  pages={132--144},
  year={2015},
  publisher={Elsevier}
}

@inproceedings{li2007fast,
  title={A fast segmentation method for stl teeth model},
  author={Li, Zhanli and Ning, Xiaojuan and Wang, Zengbo},
  booktitle={2007 IEEE/ICME International Conference on Complex Medical Engineering},
  pages={163--166},
  year={2007},
  organization={IEEE}
}

@inproceedings{qi2017pointnet,
  title={Pointnet: Deep learning on point sets for 3d classification and segmentation},
  author={Qi, Charles R and Su, Hao and Mo, Kaichun and Guibas, Leonidas J},
  booktitle={Proceedings of the IEEE conference on computer vision and pattern recognition},
  pages={652--660},
  year={2017}
}

@article{qi2017pointnet++,
  title={Pointnet++: Deep hierarchical feature learning on point sets in a metric space},
  author={Qi, Charles Ruizhongtai and Yi, Li and Su, Hao and Guibas, Leonidas J},
  journal={Advances in neural information processing systems},
  volume={30},
  year={2017}
}

@article{wang2019dynamic,
  title={Dynamic graph cnn for learning on point clouds},
  author={Wang, Yue and Sun, Yongbin and Liu, Ziwei and Sarma, Sanjay E and Bronstein, Michael M and Solomon, Justin M},
  journal={ACM Transactions on Graphics (tog)},
  volume={38},
  number={5},
  pages={1--12},
  year={2019},
  publisher={Acm New York, NY, USA}
}

@article{lian2020deep,
  title={Deep multi-scale mesh feature learning for automated labeling of raw dental surfaces from 3D intraoral scanners},
  author={Lian, Chunfeng and Wang, Li and Wu, Tai-Hsien and Wang, Fan and Yap, Pew-Thian and Ko, Ching-Chang and Shen, Dinggang},
  journal={IEEE transactions on medical imaging},
  volume={39},
  number={7},
  pages={2440--2450},
  year={2020},
  publisher={IEEE}
}

@inproceedings{xiong2023tsegformer,
  title={Tsegformer: 3d tooth segmentation in intraoral scans with geometry guided transformer},
  author={Xiong, Huimin and Li, Kunle and Tan, Kaiyuan and Feng, Yang and Zhou, Joey Tianyi and Hao, Jin and Ying, Haochao and Wu, Jian and Liu, Zuozhu},
  booktitle={International conference on medical image computing and computer-assisted intervention},
  pages={421--432},
  year={2023},
  organization={Springer}
}

@article{jin2024tsrnet,
  title={TSRNet: A Dual-Stream Network for Refining 3D Tooth Segmentation},
  author={Jin, Hairong and Shen, Yuefan and Lou, Jianwen and Zhou, Kun and Zheng, Youyi},
  journal={IEEE Transactions on Visualization and Computer Graphics},
  year={2024},
  publisher={IEEE}
}

@inproceedings{kirillov2023segment,
  title={Segment anything},
  author={Kirillov, Alexander and Mintun, Eric and Ravi, Nikhila and Mao, Hanzi and Rolland, Chloe and Gustafson, Laura and Xiao, Tete and Whitehead, Spencer and Berg, Alexander C and Lo, Wan-Yen and others},
  booktitle={Proceedings of the IEEE/CVF international conference on computer vision},
  pages={4015--4026},
  year={2023}
}

@article{ma2024segment,
  title={Segment anything in medical images},
  author={Ma, Jun and He, Yuting and Li, Feifei and Han, Lin and You, Chenyu and Wang, Bo},
  journal={Nature Communications},
  volume={15},
  number={1},
  pages={654},
  year={2024},
  publisher={Nature Publishing Group UK London}
}

@article{ke2023segment,
  title={Segment anything in high quality},
  author={Ke, Lei and Ye, Mingqiao and Danelljan, Martin and Tai, Yu-Wing and Tang, Chi-Keung and Yu, Fisher and others},
  journal={Advances in Neural Information Processing Systems},
  volume={36},
  pages={29914--29934},
  year={2023}
}

@article{fan2023stable,
  title={Stable segment anything model},
  author={Fan, Qi and Tao, Xin and Ke, Lei and Ye, Mingqiao and Zhang, Yuan and Wan, Pengfei and Wang, Zhongyuan and Tai, Yu-Wing and Tang, Chi-Keung},
  journal={arXiv preprint arXiv:2311.15776},
  year={2023}
}

@article{ravi2024sam,
  title={Sam 2: Segment anything in images and videos},
  author={Ravi, Nikhila and Gabeur, Valentin and Hu, Yuan-Ting and Hu, Ronghang and Ryali, Chaitanya and Ma, Tengyu and Khedr, Haitham and R{\"a}dle, Roman and Rolland, Chloe and Gustafson, Laura and others},
  journal={arXiv preprint arXiv:2408.00714},
  year={2024}
}

@inproceedings{xia2022vision,
  title={Vision transformer with deformable attention},
  author={Xia, Zhuofan and Pan, Xuran and Song, Shiji and Li, Li Erran and Huang, Gao},
  booktitle={Proceedings of the IEEE/CVF conference on computer vision and pattern recognition},
  pages={4794--4803},
  year={2022}
}

@inproceedings{ronneberger2015u,
  title={U-net: Convolutional networks for biomedical image segmentation},
  author={Ronneberger, Olaf and Fischer, Philipp and Brox, Thomas},
  booktitle={International Conference on Medical image computing and computer-assisted intervention},
  pages={234--241},
  year={2015},
  organization={Springer}
}

@inproceedings{boykov2001interactive,
  title={Interactive graph cuts for optimal boundary \& region segmentation of objects in ND images},
  author={Boykov, Yuri Y and Jolly, M-P},
  booktitle={Proceedings eighth IEEE international conference on computer vision. ICCV 2001},
  volume={1},
  pages={105--112},
  year={2001},
  organization={IEEE}
}

@inproceedings{carion2020end,
  title={End-to-end object detection with transformers},
  author={Carion, Nicolas and Massa, Francisco and Synnaeve, Gabriel and Usunier, Nicolas and Kirillov, Alexander and Zagoruyko, Sergey},
  booktitle={European conference on computer vision},
  pages={213--229},
  year={2020},
  organization={Springer}
}

@article{vaswani2017attention,
  title={Attention is all you need},
  author={Vaswani, Ashish and Shazeer, Noam and Parmar, Niki and Uszkoreit, Jakob and Jones, Llion and Gomez, Aidan N and Kaiser, {\L}ukasz and Polosukhin, Illia},
  journal={Advances in neural information processing systems},
  volume={30},
  year={2017}
}

@article{zheng2022teethgnn,
  title={TeethGNN: semantic 3D teeth segmentation with graph neural networks},
  author={Zheng, Youyi and Chen, Beijia and Shen, Yuefan and Shen, Kaidi},
  journal={IEEE Transactions on Visualization and Computer Graphics},
  volume={29},
  number={7},
  pages={3158--3168},
  year={2022},
  publisher={IEEE}
}

@inproceedings{zhao2021point,
  title={Point transformer},
  author={Zhao, Hengshuang and Jiang, Li and Jia, Jiaya and Torr, Philip HS and Koltun, Vladlen},
  booktitle={Proceedings of the IEEE/CVF international conference on computer vision},
  pages={16259--16268},
  year={2021}
}

@article{krenmayr2024dilatedtoothsegnet,
  title={Dilatedtoothsegnet: Tooth segmentation network on 3d dental meshes through increasing receptive vision},
  author={Krenmayr, Lucas and von Schwerin, Reinhold and Schaudt, Daniel and Riedel, Pascal and Hafner, Alexander},
  journal={Journal of Imaging Informatics in Medicine},
  volume={37},
  number={4},
  pages={1846--1862},
  year={2024},
  publisher={Springer}
}

@article{jin2025learning,
  title={Learning center-and boundary-aware instance representation for 3D tooth segmentation},
  author={Jin, Hairong and Lou, Jianwen and Lu, Zhiguo and Wu, Teng and Zhou, Kun and Zheng, Youyi},
  journal={Computers \& Graphics},
  pages={104313},
  year={2025},
  publisher={Elsevier}
}

@inproceedings{ngo2023isbnet,
  title={Isbnet: a 3d point cloud instance segmentation network with instance-aware sampling and box-aware dynamic convolution},
  author={Ngo, Tuan Duc and Hua, Binh-Son and Nguyen, Khoi},
  booktitle={Proceedings of the IEEE/CVF Conference on Computer Vision and Pattern Recognition},
  pages={13550--13559},
  year={2023}
}

@inproceedings{zhang2021tsgcnet,
  title={TSGCNet: Discriminative geometric feature learning with two-stream graph convolutional network for 3D dental model segmentation},
  author={Zhang, Lingming and Zhao, Yue and Meng, Deyu and Cui, Zhiming and Gao, Chenqiang and Gao, Xinbo and Lian, Chunfeng and Shen, Dinggang},
  booktitle={Proceedings of the IEEE/CVF Conference on Computer Vision and Pattern Recognition},
  pages={6699--6708},
  year={2021}
}

@article{cui2021tsegnet,
  title={TSegNet: An efficient and accurate tooth segmentation network on 3D dental model},
  author={Cui, Zhiming and Li, Changjian and Chen, Nenglun and Wei, Guodong and Chen, Runnan and Zhou, Yuanfeng and Shen, Dinggang and Wang, Wenping},
  journal={Medical Image Analysis},
  volume={69},
  pages={101949},
  year={2021},
  publisher={Elsevier}
}

@article{wu2022two,
  title={Two-stage mesh deep learning for automated tooth segmentation and landmark localization on 3D intraoral scans},
  author={Wu, Tai-Hsien and Lian, Chunfeng and Lee, Sanghee and Pastewait, Matthew and Piers, Christian and Liu, Jie and Wang, Fan and Wang, Li and Chiu, Chiung-Ying and Wang, Wenchi and others},
  journal={IEEE transactions on medical imaging},
  volume={41},
  number={11},
  pages={3158--3166},
  year={2022},
  publisher={IEEE}
}

@article{kuhn1955hungarian,
  title={The Hungarian method for the assignment problem},
  author={Kuhn, Harold W},
  journal={Naval research logistics quarterly},
  volume={2},
  number={1-2},
  pages={83--97},
  year={1955},
  publisher={Wiley Online Library}
}

\end{document}